\documentclass[12pt,letterpaper]{article}
\usepackage[a4paper, total={7in, 10in}]{geometry}

\usepackage{graphicx}
\graphicspath{{./}{figures/}}
\usepackage{helvet}
\usepackage{authblk}
\usepackage{hyperref}
\usepackage{amsmath} 
\usepackage{amssymb} 
\usepackage{orcidlink} 
\usepackage{adjustbox}
\usepackage{multirow}
\usepackage{makecell}
\usepackage{caption}
\usepackage{footnote}
\usepackage[super,comma,sort&compress]  
   {natbib}
\usepackage[right]{lineno} 
\usepackage{nameref}
\usepackage{xcolor}
\usepackage{amsmath}
\usepackage{amssymb}
\DeclareMathOperator*{\argmax}{argmax}
\DeclareMathOperator*{\softmax}{softmax}

\definecolor{darkgreen}{RGB}{0, 150, 0}
\newcommand{\Shada}[1]{\textcolor{black}{#1}}

\newcommand{\class}[1]{\textbf{#1}}

\makeatletter
\renewcommand{\maketitle}{\bgroup\setlength{\parindent}{0pt}
\begin{flushleft}
  \textbf{\@title}
  
  \@author
\end{flushleft}\egroup}
\makeatother

\title{Multi-label versus multi-class classification of blood cells and their aggregates in microfluidic channels}
\date{}

\author[1,2,*,\orcidlink{0000-0003-4570-8937}]{Igor Zingman}
\author[1,2, \orcidlink{0000-0002-1672-2767}]{Shada Abuhattum}
\author[1,2]{Sara Kaliman}
\author[2]{Maximilian Schl\"ogel}
\author[1,2, \orcidlink{0000-0002-5734-3130}]{Paul Müller}
\author[1,2]{Markéta Kubánková}
\author[1,2]{Nadine Ströhlein}
\author[1,2]{Manuela Hauke}
\author[1]{Lena Schnörer}
\author[1]{Martin Kräter}

\author[1,2,3]{Jochen Guck}

\affil[1]{Max Planck Institute for the Science of Light, Erlangen, Germany}
\affil[2]{Max-Planck-Zentrum f\"ur Physik und Medizin}
\affil[3]{Friedrich-Alexander Universi\"at Erlangen-N\"urnberg, Erlangen, Germany}

\affil[*]{Correspondence: igor.zingman@mpzpm.mpg.de}

\begin{document}

\maketitle

\section*{SUMMARY}
Deformability cytometry (DC) is a type of imaging flow cytometry, which uses a camera-equipped device to measure cellular stiffness in addition to other cellular properties at high throughput. Cellular properties such as area and elongation can identify cell types, but this requires prior knowledge of distinguishing properties and cannot be applied to clinically important cell aggregates. Using DC data, we evaluated conventional multi-class (MC) classification and introduced a multi-label (ML) approach for identifying blood cells and their aggregates. In particular, an ML classifier can simultaneously assign multiple cell-type labels to a single imaged event. We show that, unlike MC classification, ML classification can identify cell aggregates not represented in the training data. It also avoids the need for exhaustive, strictly defined aggregate labels, thereby simplifying and speeding up annotation. Since automated blood analyzers do not reliably analyze cell aggregates, our approach may help address this clinical gap.

\section*{KEYWORDS}
Imaging flow cytometry, Deformability cytometry, Cell aggregates,  Multi-label classification, Blood cell identification, Deep neural networks

\section*{INTRODUCTION}
\label{sec:intro}

Imaging flow cytometry~\cite{rees2022imaging} uses a flow cytometry device equipped with a camera that can capture images of cells, which allows measurement of cell properties at high throughput. In this paper, we focus on deformability cytometry~\cite{otto2015real, chen2023microfluidic}, a technology that measures cell stiffness in addition to cell morphology and texture. 
It has been widely used to characterize single blood cells, as they are the most readily available and abundant cells in suspension~\cite{toepfner2018detection}, and changes in their mechanical properties can reflect physiological changes associated with infectious diseases~\cite{toepfner2018detection, kubankova2021physical}, genetic diseases~\cite{MYH9, Chorea-Acanthocytosis}, or other conditions~\cite{Depressive_disorders, Systemic_Sclerosis, schuchardt2024omega}.

It has been shown that the appearance of particular cell aggregates in blood is also clinically important~\cite{Darras2025, mchedlishvili1993effect, popp2022circulating, krell2025protocol, cardiovascular_monocyte, cardiovascular_leukocyte, Sepsis}. For example, platelet-leukocyte aggregates are important markers for cardiovascular disease~\cite{cardiovascular_monocyte, cardiovascular_leukocyte} and sepsis~\cite{Sepsis}. Recently, platelet aggregates and platelet-monocyte aggregates were linked to COVID-19~\cite{klenk2023platelet, hottz2020platelet}. However, standard automated blood cell analyzers do not reliably identify or analyze them~\cite{klenk2023platelet}.

Reliable identification of both single cells and cell aggregates is therefore needed to isolate specific cells or aggregates of clinical interest. Cells can first be detected and segmented, which allows extraction of predefined cell properties, such as area, deformability, and brightness. These properties define a feature space in which cell types occupy characteristic regions. Single cells can then be separated and identified using the established gating procedure of \citeauthor{toepfner2018detection}, which relies on manually defined polygonal gates within this feature space. However, this approach requires prior knowledge of the cell properties that distinguish different cell types. Additionally, such gating strategies cannot readily be used to identify various cell aggregates. A further limitation of this established approach is that many single cells falling outside the predefined gates remain unclassified.

An alternative approach is to use supervised multi-class (MC) classification, which requires collection of a labeled dataset of images with different cell types (see Figure~\ref{fig:single_cell_classes}). This approach is straightforward to apply for the classification of single cells and particular doublets~\cite{krater2021aideveloper, nawaz2020intelligent, schuchardt2024omega} (e.g., red blood cell doublets), but becomes challenging when also targeting classification of a large number of possible cell combinations or aggregates. First, it is challenging to collect enough data for all possible combinations of cells, particularly because some aggregates are rare. Second, it is challenging to annotate precisely defined, mutually exclusive aggregate classes.
For example, an aggregate class defined as a single platelet (PLT) bound to a red blood cell (RBC) is not easy to rapidly distinguish from a class defined as a few PLTs bound to an RBC. To simplify and speed up annotation, some aggregate classes may instead be defined in a fuzzy way. A class may comprise a single or several PLTs bound to an RBC (see Figure~\ref{fig:aggregate_classes}).
Such fuzzy classes may overlap with other classes. A class comprising a single or several PLTs bound to an RBC is not mutually exclusive with a class comprising a single or several RBCs attached to a PLT aggregate. This is because a few PLTs may also be captured as a PLT aggregate. Overlapping classes, however, are not suitable for multi-class classification.

Here, we therefore propose to use a multi-label (ML) classification approach~\cite{boutell2004learning, bogatinovski2022comprehensive}, where each imaged event can be assigned a few labels simultaneously. In this way, the large number of all possible cell aggregates does not necessarily need to be collected and labeled for supervised training. We show that this approach allows identification of various cell aggregates that were not present in the training data. 
Moreover, the approach allows the use of training data for cell aggregates labeled with overlapping labels, which makes the labeling process easier and faster.

We implemented the standard MC approach and developed an ML approach for classifying blood cells imaged using a DC device. We used deep learning with convolutional neural networks for both approaches to build cell classifiers. For training and evaluation, we used datasets annotated using fluorescent markers. Additionally, we validated the approaches on the classification of white blood cells (WBCs) using a dataset created at different times and settings, leveraging labels obtained automatically through clustering with Gaussian mixture models (GMM)~\cite{kaliman2025automation}.

The results have shown that both MC and ML approaches are feasible and can achieve high performance. They also allow detection of special blood cell types, such as nucleated RBCs (nRBCs), and cell aggregates, which are not routinely detected by standard automated devices in clinical practice~\cite{klenk2023platelet}. Although both approaches are capable of detecting cell aggregates, only the ML approach can detect cell aggregates that are not predefined and not present in the training data. In contrast to the MC approach, the ML approach provides flexibility in the collection of the training dataset.
It allows overlapping labels and does not require creating an exhaustive set of aggregate classes that need to be detected.
It is worth noting that, although we used a deformability cytometry device in our experiments, this cell classification methodology could be more broadly applied to imaging flow cytometry.

\section*{RESULTS}
\label{sec:classification_performance}

We built classifiers using deep convolutional neural networks. First, we evaluated the standard MC classification scheme, see Figure~\ref{fig:MC_architecture}, where cell classes are strictly defined and mutually exclusive. We used the EfficientNet-B0~\cite{tan2019efficientnet} convolutional neural network to encode images into feature vectors. This neural network achieved a balanced accuracy of 97.1\% on average over the predefined cell-type classes, which is superior to other architectures we tested: LeNet~\cite{lecun1998gradient}, used by \citeauthor{krater2021aideveloper}, and ResNet~\cite{he2016deep}, used by \citeauthor{schuchardt2024omega} (see details in Table~\ref{tab:MC_performance}). Both \citeauthor{krater2021aideveloper} and \citeauthor{schuchardt2024omega} employed a similar multi-class classification approach for deformability cytometry data.

We would like to note the notation used in Table~\ref{tab:MC_performance} and subsequent tables: ``(s)'' following a cell-type name indicates that the image may contain one or more cells of that type. In some of the tables that follow, cell-type names may also be enclosed in parentheses, indicating that the presence of the corresponding cell type is optional. As described in Section~\nameref{sec:collected_data}, such flexible class definitions facilitate the collection of a labeled dataset.

Collecting a dataset with exclusive and strictly defined classes for cell aggregates is challenging, given the large number of possible cell combinations, many of which are rare. We therefore developed an alternative approach based on multi-label classification. 
This approach does not require a training dataset that exhaustively covers all aggregate classes to be detected. It also does not require mutually exclusive classes, being able to leverage vaguely defined classes. This considerably simplifies and speeds up dataset collection.

The proposed architecture for ML classification is shown in Figure~\ref{fig:ML_architecture}. A set of independent binary classifiers, or heads, allows the assignment of a few labels simultaneously to an input sample. These heads correspond to a set of inherent single-cell and aggregate classes to be predicted. Head predictions potentially allow detection of arbitrary combinations of the used cell classes. Though heads are independent neural networks, they are fed from a shared fully connected neural network, allowing the model to learn correlations between classes.
The architecture also includes a WBC multi-class classifier, which classifies mutually exclusive WBC subtypes and is restricted to samples in which a WBC was detected by the corresponding binary head. As in the case of the MC classifier, a shared convolutional neural network encodes images into feature vectors, which are fed to the subsequent multi-label and WBC classifiers. 
Since the multi-label heads form the primary, defining mechanism of this architecture, we refer to it as multi-label classification, even though the additional WBC multi-class classifier technically makes it a hybrid architecture. Further details on the architecture and its training can be found in \nameref{sec:methods_ML}.

The ML approach achieves a balanced accuracy of 97.63\% on average over its inherent classes (see details in Table~\ref{tab:ML_performance_original}). Its performance cannot be directly compared with the MC approach, because they are designed to output predictions for different sets of classes. For comparison, we therefore converted the ML predictions into mutually exclusive MC class predictions, according to the mapping presented in Table~\ref{tab:ML_to_MC}.

\subsubsection*{Both MC and ML approaches enable the identification of individual cell types and their aggregates}
\label{sec:testset_perfomance}
Here, we report the performance of the MC and ML approaches on the ground truth test set. We first evaluated the MC approach using several deep convolutional neural network architectures. Specifically, we compared LeNet~\cite{lecun1998gradient}, which has previously been used to classify single cells and red blood cell (RBC) doublets in bright-field images acquired using a deformability cytometry device~\cite{krater2021aideveloper}, with the more recent ResNet~\cite{he2016deep} and EfficientNet~\cite{tan2019efficientnet} architectures.
Our experiments confirmed that the MC approach can successfully identify both single cells and various cell aggregates. We also found that EfficientNet and ResNet achieved comparable performance and both outperformed the smaller LeNet architecture. We therefore used EfficientNet, pretrained on ImageNet~\cite{deng2009imagenet}, in all subsequent experiments because it achieved the best performance, as shown in Table~\ref{tab:MC_performance}.
As with the MC classifier, we built the ML classifier using the EfficientNet architecture. Table~\ref{tab:ML_performance_original} presents its performance on the class labels defined for the ML approach. Combinations of these labels can represent a variety of cell aggregates. 

Table~\ref{tab:MC_ML_performance} compares the performance of the MC and ML approaches. To compare the ML and MC approaches, we converted the ML predictions into the mutually exclusive classes of the MC classifier, according to the mapping presented in Table~\ref{tab:ML_to_MC}. The results show that, when both classifiers were trained using sufficient numbers of examples, they performed similarly in identifying single cells and mutually exclusive aggregate classes, although the MC approach had a slight advantage. Averaged balanced accuracy over cell classes for the MC approach reaches 97.10\% compared to 95.56\% for the ML approach.  

\begin{table}[htb]
\caption{\textbf{Comparison of the balanced accuracy performance of the multi-class and multi-label cell classification approaches on the test set}}
\label{tab:MC_ML_performance}
\renewcommand{\arraystretch}{2.2}
\begin{tabular}{|l | l | l |} 
 \hline
  \textbf{Cell type} & \makecell{\textbf{Multi-class}} & \makecell{\textbf{Multi-label} }\\ [1.0ex]
 \hline
 Basophil & \makecell{\textbf{88.02} $\pm$ 0.94 }  & \makecell{84.64 $\pm$ 0.99 }\\ [0.5ex] 
 \hline
 Eosinophil & \makecell{98.57 $\pm$ 0.05 } & \makecell{\textbf{99.05} $\pm$ 0.09 }\\ [0.5ex] 
 \hline
 Monocyte & \makecell{98.33 $\pm$ 0.12 } & \makecell{\textbf{98.37} $\pm$ 0.3 }\\ [0.5ex] 
 \hline
 Neutrophil & \makecell{96.37 $\pm$ 0.39} & \makecell{\textbf{97.57} $\pm$ 0.22 }\\ [0.5ex] 
 \hline
 Lymphocyte & \makecell{99.04 $\pm$ 0.17} & \makecell{\textbf{99.44} $\pm$ 0.08 }\\ [0.5ex] 
 \hline
 RBC & \makecell{\textbf{99.22} $\pm$ 0.02}  & \makecell{98.85 $\pm$ 0.04  }\\ [0.5ex] 
 \hline
 Nucleated RBC & \makecell{\textbf{96.57} $\pm$ 0.43} & \makecell{96.02 $\pm$ 0.36 }\\ [0.5ex]
 \hline
 Platelet & \makecell{\textbf{96.53} $\pm$ 0.28 } & \makecell{85.68 $\pm$ 0.81 }\\ [0.5ex]
 \hline
 RBC aggregate & \makecell{\textbf{99.81} $\pm$ 0.02}  & \makecell{99.77 $\pm$ 0.02 }\\ [0.5ex]
 \hline
 Platelet aggregate & \makecell{96.04 $\pm$ 0.29 } & \makecell{\textbf{97.15} $\pm$ 0.13 }\\ [0.5ex]
 \hline
 RBC \& platelet(s) & \makecell{\textbf{99.30} $\pm$ 0.07} & \makecell{98.41 $\pm$ 0.23 }\\ [0.5ex]
 \hline
 WBC \& platelet(s) & \makecell{\textbf{97.34} $\pm$ 0.09} & \makecell{91.81 $\pm$ 0.46 }\\ [0.5ex]
 \hline
 Average & \makecell{\textbf{97.10} $\pm$ 0.05} & \makecell{95.56 $\pm$ 0.12 }\\ [0.5ex]
 \hline
\end{tabular}
\newline
\caption*{Bold numbers indicate the method with higher balanced accuracy. In class names, "(s)" indicates that the class definition allows either a single cell or multiple cells of that type.}
\end{table}

\bigskip

\bigskip

\subsubsection*{Multi-label classification enables identification of cell aggregates not represented in the training data}

Here, we compare the classification performance of the MC and ML approaches on additional cell-aggregate classes for which there were no examples in the training dataset. To enable this comparison, we converted the class labels defined for the ML approach into the mutually exclusive MC classes using the mapping shown in Table~\ref{tab:ML_to_MC}. Table~\ref{tab:advantage_ML} presents the results and demonstrates an advantage of the ML approach.

The results in Table~\ref{tab:advantage_ML} show that the ML approach can identify aggregate classes that were not represented in the training data. Namely, the \class{RBC(s) \& PLT aggregate}, \class{WBC(s) \& PLT aggregate}, and \class{WBC aggregate \& (PLT(s))} classes were detected with balanced accuracies of 80.27\%, 91.58\%, and 80.83\%, respectively.
Furthermore, when the MC and ML classifiers were trained with examples of these aggregate classes, the ML classifier outperformed the MC classifier. For \class{RBC(s) \& PLT aggregate}, the ML approach achieved a balanced accuracy of 98.67\% versus 96.89\% for the MC approach, while for \class{WBC(s) \& PLT aggregate}, the ML approach achieved 95.91\% versus 90.7\% for the MC approach. This result may be explained by the overlap between the \class{RBC(s) \& PLT aggregate} and \class{RBC \& platelet(s)} classes, and between the \class{WBC(s) \& PLT aggregate} and \class{WBC \& platelet(s)} classes. Consequently, some examples may belong to two classes simultaneously, impairing the performance of the MC classifier. Unlike the MC approach, the ML approach can properly handle such overlapping classes, because its labels are not required to be mutually exclusive.
For the \class{WBC aggregate \& (PLT(s))} class, we were able to collect only a small number of examples (see Section~\nameref{sec:collected_data}), which was insufficient for training a classifier. We therefore used all the collected examples of this class for testing the ML classifier. As shown in Table~\ref{tab:advantage_ML}, the ML classifier successfully identified this aggregate class despite its absence from the training dataset.
The ability of the ML approach to identify aggregate classes not represented in the training data, however, has limitations when constituent cells undergo shape deformation upon aggregation (see Section~\nameref{Sec:limitations} for details).

\begin{table}[htb]
\caption{\textbf{Balanced accuracy on the test set for classes with and without examples in the training set.}}
\label{tab:advantage_ML}
\renewcommand{\arraystretch}{2.2}
\begin{tabular}{|l | l | l | l |} 
 \hline
 \textbf{Test class} & \makecell{\textbf{The test class} \\  \textbf{in the training set?}} & \makecell{\textbf{Multi-class}  } &  \makecell{\textbf{Multi-label} } \\ [0.5ex] 
 \hline
 \multirow{2}{*}{\makecell{RBC(s) \& PLT aggr. \\ \includegraphics[width=1.5cm]{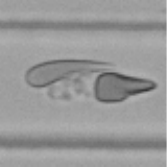}  }} & 
 yes & \makecell{96.89 $\pm$ 0.37 } & \makecell{\textbf{98.67} $\pm$ 0.27 }\\ [0.5ex] 
 \cline{2-4}
  & no & - & \makecell{80.27 $\pm$ 0.95 }\\ [0.5ex]  
 \hline
 \multirow{2}{*}{ \makecell{WBC(s) \& PLT aggr. \\ \includegraphics[width=1.5cm]{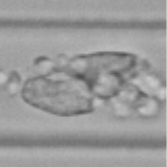} }} & yes & \makecell{90.7 $\pm$ 1.65 } & \makecell{\textbf{95.91} $\pm$ 0.73 } \\ [0.5ex]
 \cline{2-4}
 & no & - & \makecell{91.58 $\pm$ 0.97 }\\ [0.5ex] 
 \hline 
 \makecell{WBC aggr. \& (PLT(s)) \\ \includegraphics[width=1.5cm]{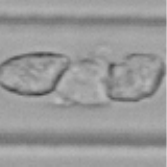}}  & no & - & \makecell{80.83 $\pm$ 1.17 }\\ [0.5ex]
 \hline 
\end{tabular}
\newline
\caption*{Bold numbers indicate the method with higher balanced accuracy. In class names, "(s)" indicates that the class definition allows either a single cell or multiple cells of that type, and names enclosed in brackets indicate that the presence of the corresponding cells is optional. Dashes indicate cases where a prediction is not possible. PLT stands for platelet.}
\end{table}
\bigskip

\subsubsection*{Multi-label classification scheme shows better generalization}
\label{sec:performance_wbc_testset}

The ground truth training and test datasets were collected from measurements belonging to different patients to avoid overfitting to patient data. However, both datasets were collected during the same time period, under similar acquisition conditions, and by the same individuals in the laboratory. We therefore collected an additional test dataset from measurements performed during a different time period and by different individuals. Sample collection followed the standard DC blood test measurement procedure, which differs slightly from the procedure used to collect the ground truth dataset. For example, the timing between blood draw and analysis may differ due to fluorescent labeling, which is required for the ground truth dataset. Fluorescent labeling may also introduce artifacts (see Section~\nameref{Sec:limitations}).

This additional test dataset was labeled using automated, unsupervised clustering of WBC features obtained from bright-field images~\cite{kaliman2025automation}. Collecting the labeled WBC test dataset required considerably less effort compared to the ground truth dataset based on fluorescent markers. It comprises a few of the most prevalent WBC subtypes: neutrophils, lymphocytes, and monocytes.

Table~\ref{tab:GMMperformance} summarizes the balanced accuracy and detection sensitivity for the WBC subtypes. Before inference, the WBC test dataset was normalized by subtracting the mean pixel value computed over the entire dataset. The results show that the proposed ML classifier generalizes better than the MC classifier to independently collected WBC data. We speculate that, because platelets were frequently attached to neutrophils in this dataset, the better performance of the ML classifier is partly attributable to its ability to identify cells (in this case, WBCs) within aggregates.

\begin{table}[htb]
\caption{\textbf{Balanced accuracy on the test set of WBCs labeled by unsupervised clustering~\cite{kaliman2025automation}. Detection sensitivity is additionally provided in brackets, below the balanced accuracy.}}
\label{tab:GMMperformance}
\renewcommand{\arraystretch}{2.2}
\begin{tabular}{|l | l | l | l | } 
 \hline
 & \textbf{Neutrophil}  & \textbf{Lymphocyte}   &  \textbf{Monocyte}  \\ [0.5ex] 
 \hline
 \textbf{Multi-class} & \makecell{ 87.36 $\pm$ 0.6  \\ (76.18 $\pm$ 1.24)} & \makecell{96.88 $\pm$ 0.3 \\ (94.19 $\pm$ 0.62)} & \makecell{ 85.86 $\pm$ 0.81 \\ (72.29 $\pm$ 1.72)}   \\ [0.5ex] 
 \hline
 \textbf{Multi-label} & \makecell{ \textbf{97.05} $\pm$ 0.17 \\ (\textbf{95.75} $\pm$ 0.39)} & \makecell{\textbf{97.92} $\pm$ 0.19 \\ (\textbf{96.35} $\pm$ 0.4)} & \makecell{ \textbf{89.96} $\pm$ 0.77  \\ (\textbf{81.95} $\pm$ 1.51)}  \\ [0.5ex] 
 \hline 
\end{tabular}
\newline
\caption*{The best results are depicted in bold.}
\end{table}

\section*{DISCUSSION}
\label{Sec:discussion}

In our work, we evaluated the feasibility of automatic classification of cell aggregates as well as single cells in image data obtained using a deformability cytometry device. In contrast to established approaches that cluster predefined morphological and textural cell features followed by cluster identification, classification approaches do not rely on predefined features. Such features are not always easily identifiable. Instead, cell-distinctive features are automatically learned during training of the classifiers. This is particularly advantageous for cell types that are not easily distinguishable from others. Additionally, cell detection based on predefined features might miss cells that are attached to other cells, forming an aggregate. Different types of cell aggregates might be hard to recognize using predefined morphological or textural features.
As mentioned above, cell aggregates are clinically important, yet automated blood cell analyzers do not reliably analyze them.

We developed an approach for cell-aggregate and single-cell classification based on multi-label principles. Unlike MC classification, this approach does not require mutually exclusive classes to be defined, which simplifies annotation of the training dataset. It also does not require collecting examples from all possible aggregate classes, while still being able to detect certain aggregate types that were not present in the training data. For example, the ML approach can detect the appearance of an RBC attached to a platelet aggregate, even if it was not trained on this specific combination, but only on single RBCs, platelet aggregates, and possibly their combinations with other cells. It can also detect a particular WBC subtype within a WBC aggregate, even when trained only on WBC aggregates without subtype labels and on individually labeled single WBC subtypes. In addition, detection of anomalies or artifacts can easily be incorporated into the ML classification scheme. Such events can be identified when all ML binary classifiers output low probability values. Another advantage of the ML approach, which we found experimentally using the WBC test dataset, is its better generalization performance.

The classification methods studied here allow detection of cell aggregates of various types, which are not reliably identified or characterized by automated hematology analyzers. These aggregates are usually ignored or flagged as abnormal in standard clinical blood analysis practice, which also lowers the reported counts for cells within an aggregate. Moreover, these methods are label-free (or stain-free), which makes them considerably cheaper for blood cell analysis compared with established fluorescent flow cytometry methods that use fluorescent markers. We therefore believe that our feasibility study, conducted using a deformability cytometry device, will promote the advancement of various imaging-based technologies as attractive, inexpensive potential alternatives to established methods for clinical blood analysis.
 
\subsection*{Limitations}
\label{Sec:limitations}
A cell that is attached to another cell within an aggregate may change its shape. 
If this deformation is substantial and not represented in the training set, it can be difficult for the ML approach to detect such aggregate types.
This might be the case when larger cells form an aggregate but appear only as single, undeformed cells in the training set. This is illustrated in Table~\ref{tab:disadvanatge_ML}. 
When the \class{RBC(s) \& WBC(s) \& (PLT(s))} aggregate type was not included in the training dataset, the ML system failed to detect it, achieving a balanced accuracy of only 51.73\%. Once this aggregate, containing cells with deformed shapes, was included in the training data, its detection became possible, with a balanced accuracy of 88.77\%. Thus, for the ML approach to successfully detect aggregates composed of cells that may be deformed when bound together, corresponding examples must still be included in the training data.
Note that, once samples from this class were included in the training set, the MC approach achieved a balanced accuracy of 99.76\%, outperforming the ML approach. We can explain this result by the fact that, for each cell type, the ML training data combines both deformed cells within aggregates and undeformed single cells, whereas the MC training data for this specific class contains only deformed cells within the aggregate.

Table~\ref{tab:MC_ML_performance} reports balanced accuracy based on ground truth labeled using fluorescent markers. 
Table~\ref{tab:GMMperformance} reports balanced accuracy for detection of the most prevalent WBC cell types, where labels were based on automated, unsupervised clustering~\cite{kaliman2025automation}. Although balanced accuracy in the latter case was computed only for a subset of cell types, we expect that the balanced accuracy for WBCs is comparable across these datasets\footnote{Since the WBC test dataset contains only a few classes, the number of negative examples for each binary problem, e.g., monocyte detection, is much lower than in the ground truth dataset labeled using fluorescent markers. This may affect detection specificity and, consequently, balanced accuracy.}. We observe, however, somewhat lower performance on the WBC test dataset, especially for monocytes and, for the MC approach, also for neutrophils. One possible reason is moderate overfitting of the neural networks to the specific blood measurement conditions. The WBC test dataset was collected at different times, by different individuals, and under different device settings. Collecting a larger labeled ground truth dataset with cells captured under more diverse acquisition conditions would likely increase robustness to these factors. 

Another, more fundamental reason for the lowered performance on some WBC subtypes may be related to the fluorescent labeling procedure itself. We noticed that fluorescently labeled WBCs frequently appear somewhat different, e.g., with a disrupted cell membrane. We suspect that fluorescent labeling sometimes caused cells to become activated. Thus, the training data may contain features that do not appear in real test data, such as the WBC test dataset used to report performance in Table~\ref{tab:GMMperformance}. This effect was most noticeable for monocytes.

\begin{table}[htb]
\caption{\textbf{Balanced accuracy on the test set for cell aggregates in which attached cells undergo shape deformation}}
\label{tab:disadvanatge_ML}
\renewcommand{\arraystretch}{2.2}
\begin{tabular}{|l | l | l | l |} 
 \hline
 \textbf{Test class} & \makecell{\textbf{The test class} \\ \textbf{in the training set?}} & \makecell{\textbf{Multi-class} } &  \makecell{\textbf{Multi-label}} \\ [0.5ex] 
 \hline
 \multirow{2}{*}{
 \makecell{RBC(s) \& WBC(s) \& (PLT(s)) \\ \includegraphics[width=1.5cm]{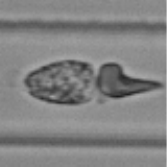} }} & yes & \makecell{\textbf{99.76} $\pm$ 0.06  } & \makecell{88.77 $\pm$ 2.86 }\\ [0.5ex] 
 \cline{2-4}
  & no & - & \makecell{51.73 $\pm$ 0.54 }\\ [0.5ex]   
 \hline 
\end{tabular}
\newline
\caption*{Bold numbers indicate the method with higher balanced accuracy. In class names, "(s)" indicates that the class definition allows either a single cell or multiple cells of that type, and names enclosed in brackets indicate that the presence of the corresponding cells is optional. Dashes indicate cases where a prediction is not possible.}
\end{table}

\newpage

\section*{METHODS}


\subsection*{Deformability cytometry and blood measurement protocol}

Deformability cytometry (DC) is a microfluidic technique for measuring the mechanical properties of single cells at high throughput~\cite{otto2015real}. 
As cells pass through a microfluidic channel of DC device, they deform under the shear stress imposed by a highly viscous surrounding fluid containing methyl-cellulose (MC). The region of interest (ROI) is placed at the end of the channel where cells reach steady state deformation and are imaged in bright field. If a fluorescence module is used, excitation lasers are also focused on cells in ROI and the emitted signal is detected in a multi-channel detector array. 

Venous blood samples collected in citrate or EDTA tubes were analyzed with a commercial DC instrument (AcCellerator, Zellmechanik Dresden GmbH), following a previously established protocol~\cite{toepfner2018detection, kaliman2025automation}. Blood donors signed written informed consent in agreement with the Declaration of Helsinki, under the permission of the local ethics committee of the Faculty of Medicine, Friedrich-Alexander-University Erlangen-Nuremberg (no. 295-20-B and no. 13\_21 Bc).

The PDMS-based microfluidic chips contain a narrow constriction channel with 20 $\times$ 20 $\mu$m cross-section and are redesigned to have 500 $\mu$m in length. PDMS chips are perfused with sample and sheath streams at a flow rate ratio of 1:3. For our experiments we used combined flow rate of 0.06 $\mu$L/s. While sheath tube contains only MB, the sample tube contains blood sample mixed with measurement buffer (MB; 0.59\% MC in phosphate-buffered saline) with ratio 1:19. We used 10 $\mu$L of whole blood which was mixed well with MB immediately  drawn via a syringe pump in reverse-flow mode into sample tubing preloaded with MB. Cell images were captured in bright field at 250 $\times$ 80 pixel resolution, recorded at 3600 frames per second over a 10-minute acquisition window.

\subsection*{Ground truth dataset annotated using fluorescent markers}
\label{sec:collected_data}
 Bright-field images for the training and test datasets were captured using a deformability cytometry device (AcCellerator, Zellmechanik Dresden GmbH) from blood measurements performed by different individuals to reduce human bias. RBCs were labeled manually, as they are visually distinctive and can be relatively easily identified by eye. All other cell types were annotated using fluorescent markers~\cite{rosendahl2018real}. 
 
 \Shada{White blood cell subtypes were identified using fluorescently conjugated antibodies against lineage-specific surface markers, with the corresponding fluorescence channels recorded simultaneously with the bright-field image for each event. Lymphocyte subsets were resolved with a CD19 / CD3 / CD56 panel as B cells (CD19+, CD3$-$, CD56$-$), T cells (CD3+, CD19$-$, CD56$-$), and NK cells (CD56+, CD3$-$). Eosinophils and basophils were identified as double-positive populations, CD193 (CCR3)+ / Siglec-8+ and CD123+ / CD203c+, respectively. Neutrophils were identified using either of two marker panels, CD15+ / CD49d$-$ / CD16+ or CD16+ / CD66+ / CD14$-$. Monocytes were identified either positively, as CD14+ / CD66$-$ / CD16$-$, or as the fraction negative for all markers of the CD19 / CD3 / CD56 / CD66 panel (CD19$-$, CD3$-$, CD56$-$, CD66$-$). Platelets were labeled using a CD61 / CD62P / CD144 panel, covering platelet identity and activation markers. Samples were pre-incubated with an Fc-receptor blocking reagent, then incubated with the respective antibody panel for 20 min at room temperature in the dark. Nucleated red blood cells (nRBCs) were resolved as cells with positive nuclear staining and negative pan-leukocyte surface marker, Draq5+ / CD45-.}

\Shada{Cell labeling was performed with an in-house annotation tool~\cite{dctag}. For each event, the tool displays the segmented cell within a bounding box together with its measured features, including deformation, brightness, and the corresponding fluorescence-channel signals (when present). The annotator uses this combined morphological and fluorescence information to confirm that the event is the cell type of interest, and then accepts or assigns the appropriate label. To make annotation efficient, the tool operates on files that were pre-enriched for the target cell type, so that each measurement contains a high proportion of the cells of interest.}

\begin{figure}
\noindent\includegraphics[width=1.0\linewidth]{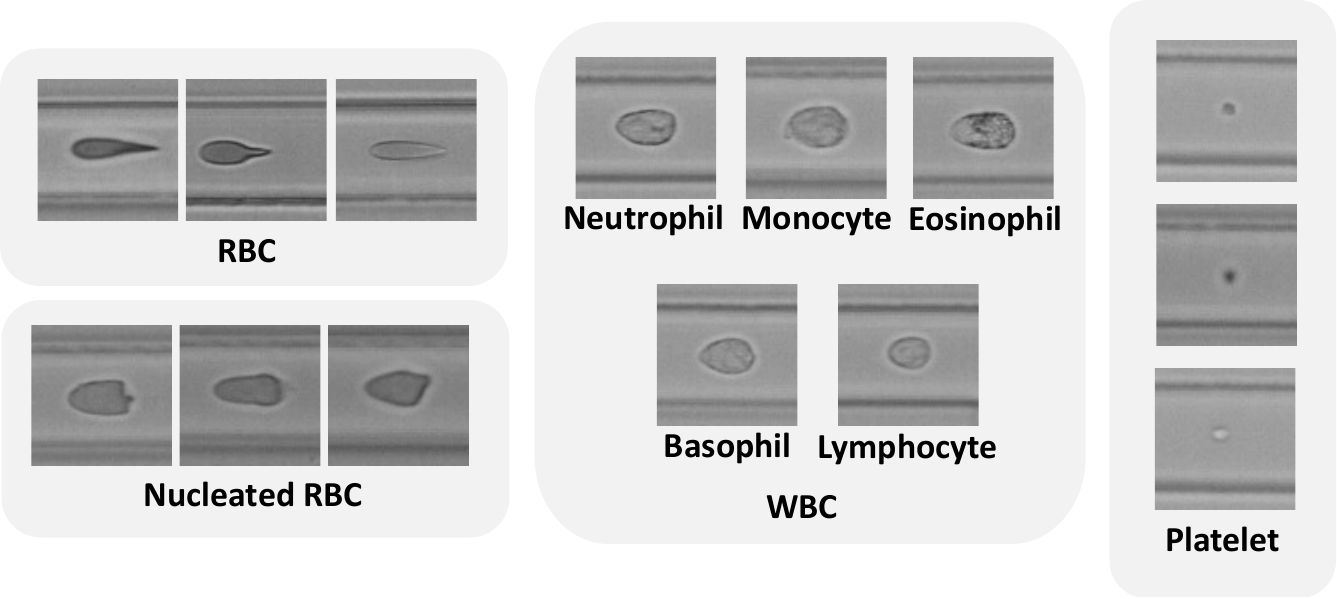}
\caption{Single-cell classes. RBC stands for Red Blood Cell, also known as erythrocyte. WBC stands for White Blood Cell, also known as leukocyte.}
\label{fig:single_cell_classes}
\end{figure}

\begin{figure}
\noindent\includegraphics[width=1.0\linewidth]{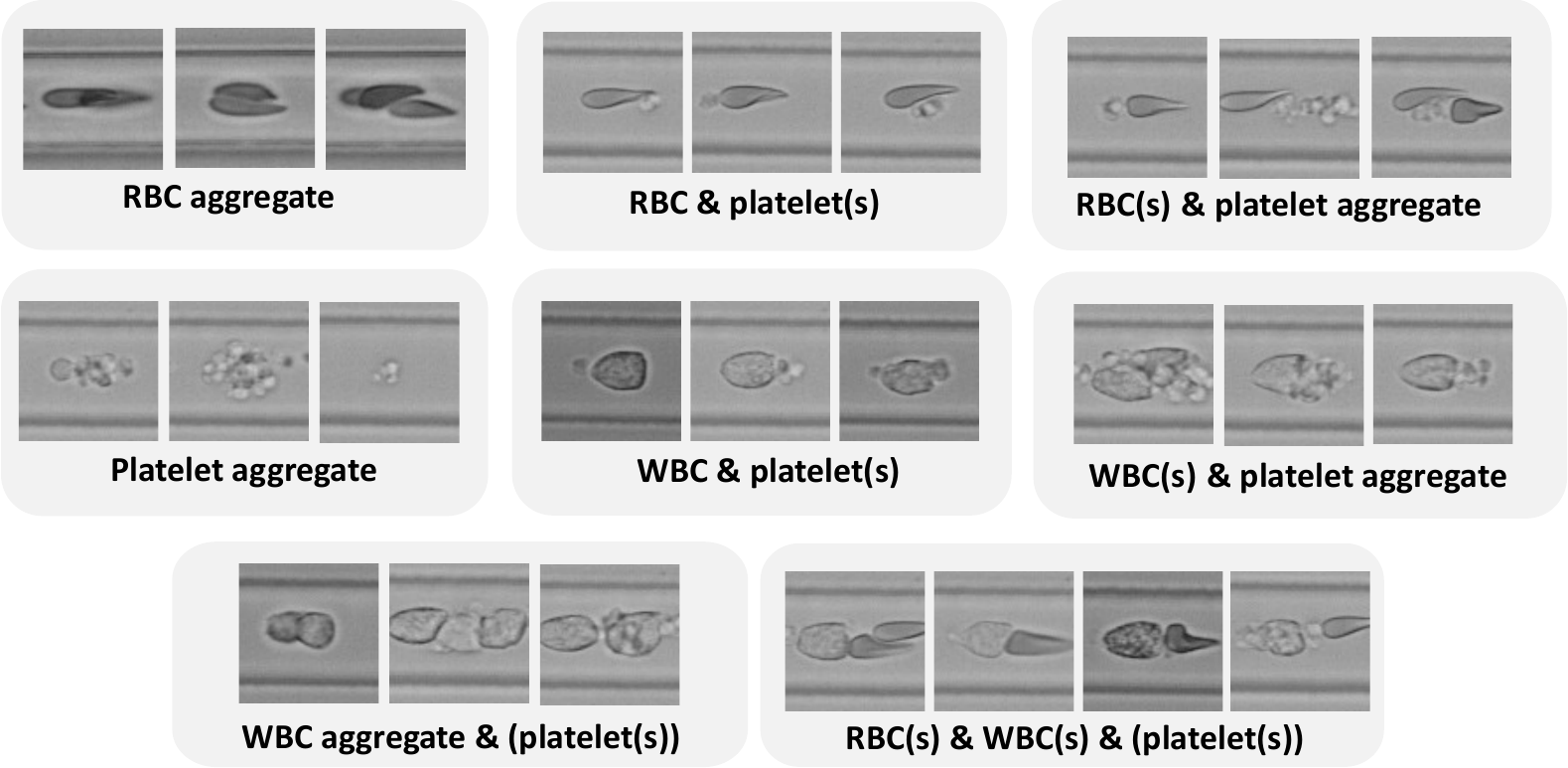}
\caption{Aggregate classes. A cell name in brackets indicates that the corresponding cell type may be present but is not required for the image to belong to the class. An "(s)" following a cell name indicates that the image may contain a single cell or several cells of that type.}
\label{fig:aggregate_classes}
\end{figure}

Examples of labeled classes for single cells and aggregates are shown in Figures~\ref{fig:single_cell_classes} and \ref{fig:aggregate_classes}, respectively. We collected a dataset containing over 122,000 training and over 42,000 test examples.
The training dataset contains 52,899 RBCs, 8,022 nucleated red blood cells (nRBCs), 15,601 WBCs, and 5,609 single platelets, as well as 25,273 RBC aggregates and 14,666 other cell aggregates. The dataset is highly unbalanced: the proportions of different cell types vary considerably due to differences in their natural abundance in the human body. For example, only 199 basophils (a type of WBC that is very scarce in blood) were collected for the training dataset. The training dataset was divided into training and validation subsets, comprising 80\% and 20\% of the cell images, respectively. The validation subset was used to select the best-performing neural network within, see details in section \ref{sec:training}.

The test dataset was collected as part of the same effort and contains 20,890 RBCs, 1,566 nucleated RBCs, 1,703 WBCs, and 1,679 single platelets, as well as 12,334 RBC aggregates and 4,537 other cell aggregates. We were able to collect only small numbers of basophils and WBC aggregates: 66 basophils and 104 WBC aggregates were collected for the test dataset. The training dataset did not include WBC aggregates. We use WBC aggregates to demonstrate that the ML approach can still identify this category, even without being trained on it.

Since we aimed to identify not only individual cells (Figure~\ref{fig:single_cell_classes}) but also various cell aggregates (Figure~\ref{fig:aggregate_classes}), defining exclusive classes precisely becomes challenging. For example, an aggregate class defined to include an RBC and a single platelet is difficult to label rapidly, either manually or using fluorescent labels. Determining whether an image containing just a few platelets belongs to a class with a single platelet or a platelet aggregate is difficult and time-consuming. We therefore defined some classes in a fuzzy way, as shown in Figure~\ref{fig:aggregate_classes}.

\subsection*{WBC test dataset generated by automated clustering}
\label{sec:GMM dataset}

In addition to the dataset with ground-truth labels derived from fluorescence measurements, we built an additional labeled test dataset based on automated clustering of WBCs~\cite{kaliman2025automation}, which required considerably less annotation effort. It was collected at a different time and by different individuals than the training dataset, and is therefore expected to have a different data distribution. This dataset comprises the most prevalent WBC subtypes: 1,542 neutrophils, 169 monocytes, and 539 lymphocytes. 

\subsection*{Evaluation metrics}

The number of examples for different cell types and cell-aggregate types in the collected dataset varies substantially. Furthermore, in the multi-label approach, the classifier is designed and trained to solve a set of binary problems (see Figure~\ref{fig:ML_architecture}) in a one-vs-rest formulation. In each binary problem, the number of positive examples is much smaller than the number of negative examples, because the negative class consists of examples from all other classes. We therefore need a performance measure that is appropriate for heavily imbalanced classes. Common choices in this setting include the F1 score and balanced accuracy.
The F1 score is more informative than ordinary accuracy when the dataset is imbalanced, because accuracy can be high even if the classifier performs well only on the prevalent class while performing poorly on rare classes. The F1 score balances precision and recall (sensitivity) for the positive class. However, F1 does not account for true negatives and therefore does not reflect performance on the negative class, i.e., specificity. Moreover, F1 can be affected by class prevalence: when the positive class is rare and negative examples are highly prevalent, even a small false-positive rate can substantially reduce precision, and therefore F1~\cite{hays2023simplistic}.

We therefore prefer to use balanced accuracy, which is the mean of sensitivity and specificity and is thus less sensitive to class prevalence than ordinary accuracy or F1. This property is especially important in our setting with multiple classes, where we compute overall performance by averaging class-specific performances, and each class may exhibit a different degree of positive-negative imbalance.

In Section~\nameref{sec:classification_performance}, we compute standard errors for performance measures $S$ (such as balanced accuracy and sensitivity) using multi-seed evaluation. We trained an ensemble of six neural networks using a set of seeds (7, 17, 27, 37, 47, 57), which, among other things, initialize the networks with different weights and feed the training data in a different order. For each trained network in the ensemble (i.e., for each seed), we calculate the classification performance for each cell or cell-aggregate type, $S_{cell, \, seed}$, as well as the average performance across all cell classes, $\bar{S}_{seed}$. For each cell class, we report the average performance across all trained networks, $\bar{S}_{cell}$, along with the corresponding standard error. We also report the average of $\bar{S}_{seed}$ across all seeds, along with the corresponding standard error, as a representative performance measure, $\bar{S}$, of the evaluated method.

\subsection*{Multi-class cell classification}
\label{sec:methods_MC}

\begin{figure}[htbp]
\noindent\includegraphics[width=1.0\linewidth]{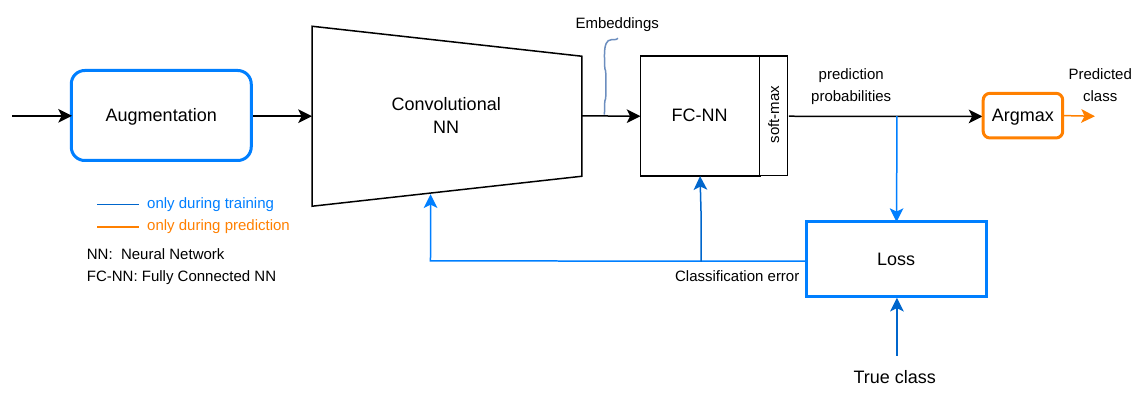}
\caption{Multi-class cell classification architecture}
\label{fig:MC_architecture}
\end{figure}

We implemented the standard multi-class (MC) classification approach according to the scheme shown in Figure~\ref{fig:MC_architecture}. Cells are classified into one of $N$ exclusive classes. 
In the experiments comparing MC and ML classification, summarized in Table~\ref{tab:MC_ML_performance}, $N$ is equal to twelve.
We used an EfficientNet-B0~\cite{tan2019efficientnet} convolutional neural network, pretrained on ImageNet~\cite{deng2009imagenet}, to generate image feature vectors. This architecture outperformed the other architectures we tested, LeNet~\cite{lecun1998gradient} and ResNet~\cite{he2016deep} (see Results, Section~\nameref{sec:classification_performance}, for details).
We truncated the last three mobile inverted bottleneck (MBConv) stages of EfficientNet, leaving only 80 channels at the output of the convolutional neural network. The resulting 80-dimensional feature vectors were then fed into a one-layer fully connected neural network, followed by a $\softmax$ non-linearity and an $\argmax$ function for the final classification into one of the exclusive cell classes. The first convolutional layer of the pretrained EfficientNet has three-channel filters for processing RGB images. Since our data consist of gray-tone images, we replaced the first convolutional layer with one-channel filters, whose weights were initialized by averaging the weights of the pretrained three-channel filters.
The neural network classifier was fed cropped $80 \times 40$ cell images. We cropped the gray-tone images around the center of mass of the binary mask of the segmented cells~\cite{U-Net}.

We used the cross-entropy loss function to train the neural network. Our dataset has highly unbalanced classes, with very different numbers of samples per class. Therefore, during training, we normalized the number of examples per class through balanced sampling. This technique oversamples rare classes (each sample is still different due to random augmentations) and undersamples frequent classes. The resulting, uniformly sampled data helps prevent the network from ignoring rare classes.

\subsection*{Multi-label cell classification}
\label{sec:methods_ML}

We developed a multi-label (ML) classification~\cite{boutell2004learning, bogatinovski2022comprehensive} approach for classifying cells and cell aggregates in a DC device. In this approach, each imaged event can be assigned to a few classes simultaneously. For example, an event containing a RBC bound to a platelet can be assigned to both the RBC and platelet classes, thereby detecting a cell aggregate. Note that, in contrast to multi-label classification, multi-class classification requires a specific aggregate category, RBC with a platelet, to be predefined and used for training in order to properly recognize such an event. The advantages of multi-label classification for cell detection in DC are discussed in the \nameref{sec:intro}, \nameref{sec:classification_performance}, and \nameref{Sec:discussion} sections. 

\begin{figure}
\noindent\includegraphics[width=1.05\linewidth]{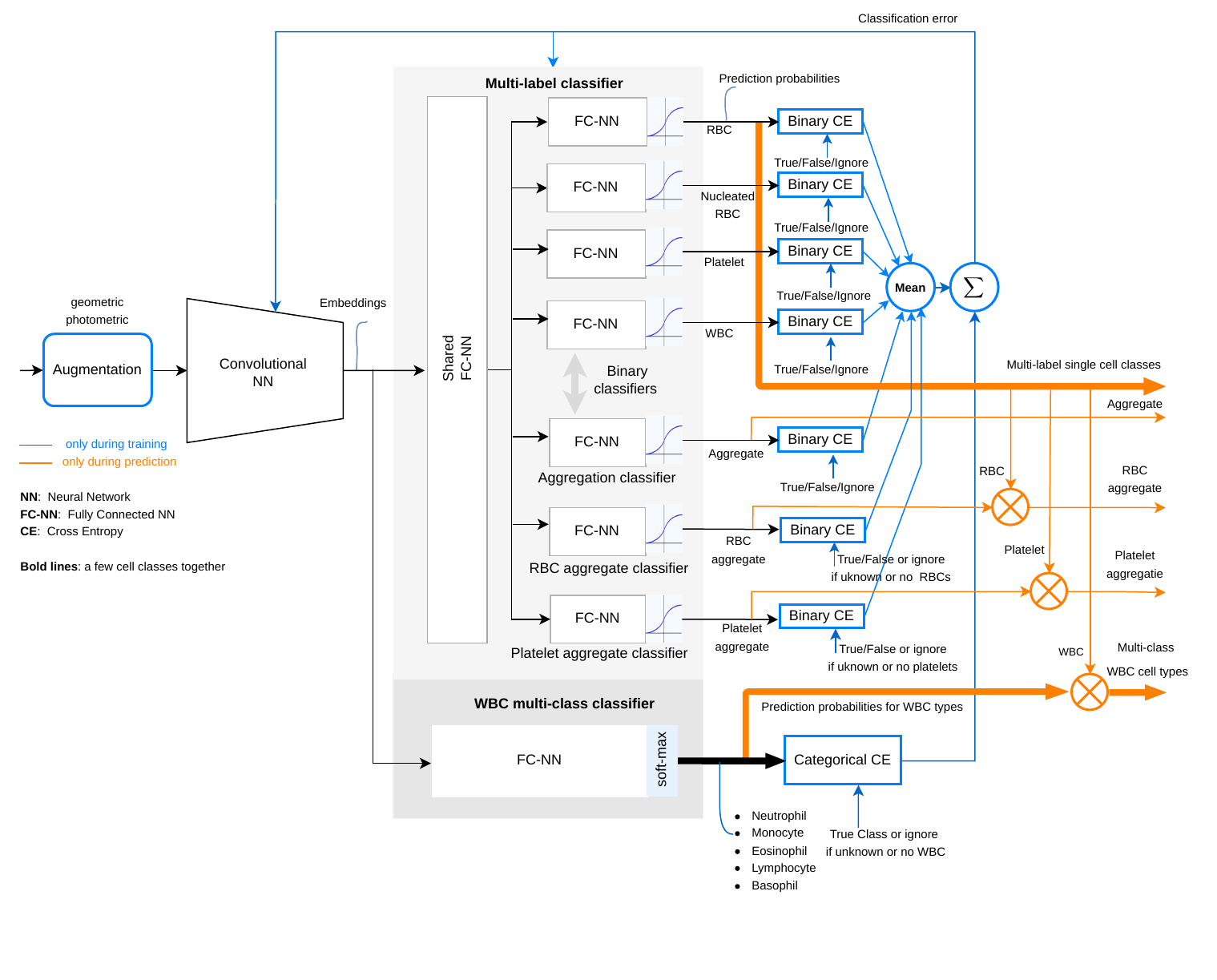}
\caption{Proposed multi-label cell classification architecture. Note that multi-class classification of WBC subtypes is part of the architecture.}
\label{fig:ML_architecture}
\end{figure}

Our multi-label cell classification algorithm is illustrated in Figure~\ref{fig:ML_architecture}. It consists of two parts: a multi-label classifier and a WBC multi-class classifier.
\footnote{This architecture could be called hybrid; however, we refer to it as multi-label, since the multi-class classifier is a sub-component refining WBC predictions within the broader multi-label architecture.}
The latter is intended solely for classifying white blood cells into one of several exclusive WBC subtypes. The former, the multi-label classifier, is composed of a set of $N=7$ binary classifiers designed to identify RBCs, nRBCs, platelets, WBCs (of any subtype), and a few types of cell aggregates. Specifically, we designed binary classifiers for RBC aggregates, platelet aggregates, and a generic aggregate class that identifies any type of aggregate. We did not design a WBC aggregate classifier, solely because we did not have sufficient data to train a classifier for this aggregate type. 

The binary classifiers of the multi-label classifier are independent heads, each implemented as a two-layer fully connected neural network followed by a sigmoid non-linearity, and can simultaneously assign high probabilities to scenes that include several cell types. In each head, both the input and hidden layers have a number of neurons equal to $N$. All heads are fed from a shared, one-layer fully connected neural network with $N$ output neurons. Although we designed independent heads, each corresponding to a specific cell or cell-aggregate type, the preceding shared neural network allows the model to learn correlations between different classes.

The WBC multi-class classifier is implemented as a one-layer fully connected neural network, which outputs a score for each WBC cell type, followed by a $\softmax$ non-linearity that converts these scores into probabilities, emphasizing the highest score. Both the multi-label and WBC multi-class classifiers share the same input feature vectors (embeddings), generated from images by a deep convolutional neural network. Specifically, we used the same modified and truncated EfficientNet-B0 architecture described in the previous \nameref{sec:methods_MC} section, which takes gray-tone images as input and outputs 80-dimensional feature vectors.

We trained the multi-label and WBC multi-class classifiers simultaneously on the dataset described in Section~\nameref{sec:collected_data}. The weights of the convolutional neural network, the shared fully connected network and binary classifier heads of the multi-label classifier, and the fully connected network of the WBC multi-class classifier were all jointly optimized.
The WBC multi-class classifier was trained using the categorical cross-entropy loss function, $CE_{MC}$. Only image samples $x_i$ labeled with one of the WBC subtypes $\mathcal{W}_j$ were considered; all other samples were excluded from the loss term:
\begin{equation}
    CE_{MC} = - \sum_{x_i \in \mathcal{W}} \sum_{j=1}^M y_{x_i}^j \log P_j(x_i),
\end{equation}
where $P_j(x_i)$ denotes the predicted probability that sample $x_i$ belongs to WBC subtype $\mathcal{W}_j$, $\mathcal{W} = \bigcup_{j=1}^{M} \mathcal{W}_j$, $M$ is the number of WBC subtypes ($M=5$ in our case), and the true label is defined as
\begin{equation*}
    y_{x_i}^j = \begin{cases}
            1, & \text{if } x_i \in \mathcal{W}_j, \\
            0, & \text{otherwise}.
        \end{cases}
\end{equation*}

\noindent The multi-label classifier was trained using a loss function, $CE_{ML}$, defined as the average of the binary cross-entropy loss functions $CE_{ML}^j$ for each of the corresponding binary classifiers $j$ (defined below), with equal weights:
\begin{equation}
CE_{ML} = \frac{1}{N}\sum_{j=1}^N CE_{ML}^j,
\end{equation}
where $N$ is the number of binary classifiers. We trained the binary classifiers in a one-versus-all manner. Positive samples were all samples containing a cell or cell aggregate from the target class, while negative samples were drawn from all other classes. Samples with vaguely defined cells were excluded from the corresponding loss function. For example, samples labeled as \class{RBC \& platelet(s)} were entirely excluded when training the \class{platelet aggregate} classifier, because such samples may or may not contain a platelet aggregate. In this way, we make use of fuzzily labeled samples when training each binary classifier, relying only on the cells or cell aggregates that are confidently labeled within the scene and ignoring cells with indefinite labels.

As in the case of multi-class classification, described in the previous section, we employ balanced sampling to ensure a similar number of examples when training the WBC multi-class classifier, which classifies single WBCs into subtypes. However, when training the binary classifiers of the multi-label classifier using the one-versus-all strategy, the number of negative examples is unknown a priori and is typically much higher than the number of positive examples. Therefore, for each binary problem $j$, we first compute the prevalence of the negative class, i.e., the ratio of the number of negative examples to the number of positive examples, and then use this ratio during training as a positive weight $W^+_j$ within the binary cross-entropy loss, in order to amplify the importance of the positive class:
\begin{equation}
CE_{ML}^j = -\sum_{x_i \in \mathcal{P}_j \cup \mathcal{N}_j} \left[ W^+_j\, y_{x_i}^j \log P_j(x_i) + (1-y_{x_i}^j) \log\left(1-P_j(x_i)\right) \right],
\end{equation}
where $P_j(x_i)$ is the positive class (target) probability estimated by binary classifier $j$ for input sample $x_i$, $\mathcal{P}_j$ and $\mathcal{N}_j$ are the sets of positive and negative samples, respectively, for the binary classification task $j$, and the true labels are defined as
\begin{equation*}
y_{x_i}^j = \begin{cases}
    1, & \text{if } x_i \in \mathcal{P}_j, \\
    0, & \text{if } x_i \in \mathcal{N}_j.
\end{cases}
\end{equation*}

\noindent For the final loss $\mathcal{L}$, we assign equal weight to the categorical cross-entropy loss of the WBC multi-class classifier, $CE_{MC}$, and the average binary cross-entropy loss of the multi-label classifier, $CE_{ML}$:
\begin{equation}
\mathcal{L} = CE_{MC} + CE_{ML}.
\end{equation}

\noindent Note that the aggregate classes of the multi-label classifier are not independent of the single-cell classes. For example, the RBC aggregate class should be detected only if the RBC single-cell class is detected by a separate, corresponding binary classifier; the platelet aggregate class should be detected only if the platelet single-cell class is detected; and the generic aggregate class should be detected only if more than one single-cell class is detected. Although the corresponding binary heads of the multi-label classifier are trained independently, we expect them to learn this dependency from the relationships between the training subsets used for each binary classifier. For example, the RBC aggregate classifier is trained on a subset of the data used to train the RBC classifier: the RBC training data includes samples containing either a single RBC or an RBC aggregate, whereas the training data for the RBC aggregate classifier includes only samples containing an RBC aggregate, and no samples containing a single RBC cell. However, because there is no strict guarantee that the independent aggregate heads will output a high score only when the corresponding single-cell classifiers also output a high score, we constrained the aggregate heads' outputs by multiplying them with the outputs of the corresponding single-cell heads, as shown in Figure~\ref{fig:ML_architecture}. Additionally, we constrained the WBC multi-class classifier by multiplying its probabilities with the output of the binary WBC classifier. This was necessary because the WBC multi-class classifier was trained only on samples containing WBCs, and can therefore produce unpredictable results for samples containing other cell types.

\begin{figure}
\noindent\includegraphics[width=0.5\linewidth]{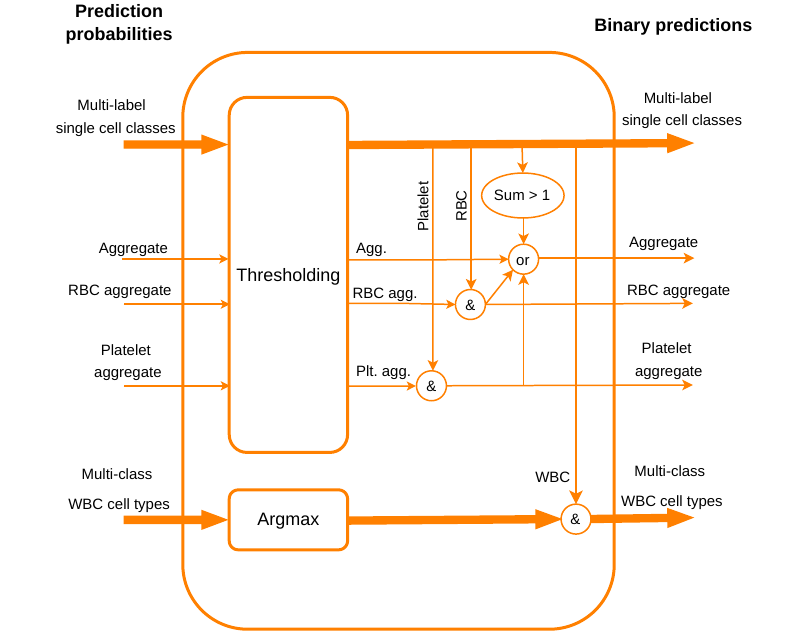}
\caption{Binary predictions for the multi-label classification architecture}
\label{fig:binary_decision}
\end{figure}

Binary decisions at inference time are made by thresholding the multi-label predicted probabilities (with a threshold of 0.5) and by applying the $\argmax$ function to the multi-class predicted probabilities, according to the scheme depicted in Figure~\ref{fig:binary_decision}. Similar to constraining the probabilities as described above, we also constrained the binary decisions for cell aggregates and WBC subtypes on the binary decisions for the single-cell outputs of the multi-label classifier, which is also shown in Figure~\ref{fig:binary_decision}.

\subsection*{Training procedure}
\label{sec:training}

Models were optimized by maximizing balanced accuracy on the validation set. Each model was trained for 20 epochs, and the checkpoint with the highest validation balanced accuracy was selected. We used the Adam optimizer~\cite{kingma2014adam} with an initial learning rate of $10^{-4}$, which was multiplied by $\gamma=0.1$ every six epochs. Input images were augmented using random flips, photometric transformations (gamma correction and adjustments to brightness and contrast), image corruptions (Gaussian noise and blurring), and color inversion. The transformation parameters were sampled independently for each augmented image. The dataset was normalized to have zero mean and unit variance.

\newpage


\section*{RESOURCE AVAILABILITY}


\subsection*{Lead contact}


Requests for further information and resources should be directed to and will be fulfilled by the lead contact, Igor Zingman (igor.zingman@mpzpm.mpg.de).

\subsection*{Data and code availability}


\begin{itemize}
    \item The DC dataset of imaged and labeled cells used for training, validation, and testing, as well as the WBC test dataset, labeled using automated, unsupervised clustering~\cite{kaliman2025automation} are publicly available at \url{https://osf.io/3zkvw}.    
    \item The training and evaluation code is publicly available in a GitHub repository \url{https://github.com/GuckLab/Single-and-aggregate-cell-classification.git}. 
    \item Any additional information required to reanalyze the data reported in this paper is available from the lead contact upon request.
\end{itemize}

\section*{ACKNOWLEDGMENTS}


We would like to thank Nadia Sbaa for her valuable contribution to data collection and implementation of the MC approach, Eoghan O'Connell for his help in building the annotation tool, and Raghava Alajangi for helpful discussions about the methods used. 
The authors are also grateful to Prof. Manfred Rauh and Prof. Markus Metzler for their support in obtaining samples from the Pediatrics Department at University Hospital Erlangen. The authors acknowledge financial support through the Medical Valley Award and BayVFP program provided by the Bavarian State Ministry of Economic Affairs, Regional Development and Energy (LSM-2303-0012), as well as core funding from the Max Planck Society. Marketa Kubankova is grateful for support via the Hermann Neuhaus Prize from the Max Planck Society funded by the Max Planck Foundation.
Microfluidic chips used in this study were kindly provided by Parth Patel and Salvatore Girardo (Core Facility Lab-on-a-Chip, Max-Planck-Zentrum für Physik und Medizin).
The authors thank all members of the lab for their support.

\section*{AUTHOR CONTRIBUTIONS}


M.Kräter, S.A., S.K., J.G. conceived the initial project;
P.M. designed the annotation tool;
S.A., S.K., M.Kubánková, N.S., M.H., L.S. collected, annotated, and curated the data;
M.S. initiated the implementation of the MC approach;
I.Z. revised the MC approach, proposed and implemented the ML approach, conceptualized the manuscript, and analyzed the data;
I.Z. wrote the manuscript with contributions from S.A., S.K., M.Kubánková;
S.A., S.K., M.Kubánková, M.Kräter, P.M. revised the manuscript;
S.A., S.K. supervised the project;
J.G. provided resources and funding.

\section*{DECLARATION OF INTERESTS}


S.A., P.M., and M.Kräter are co-founders and shareholders of the company Rivercyte GmbH, which commercializes products for DC analysis of blood samples. P.M., and M.Kräter are currently affiliated with Rivercyte GmbH.
The other authors declare no conflict of interest.

\section*{DECLARATION OF GENERATIVE AI AND AI-ASSISTED TECHNOLOGIES IN THE MANUSCRIPT PREPARATION PROCESS}


During the preparation of this work, the authors used an LLM (Antropic Claude Sonnet-5) for improving the language clarity. The authors reviewed and edited the output as needed and take full responsibility for the content of the published article. 

\newpage





\newpage


\bibliography{references}


\clearpage

\section*{Supplemental information}

\setcounter{figure}{0}
\renewcommand{\thefigure}{S\arabic{figure}}

\setcounter{table}{0}
\renewcommand{\thetable}{S\arabic{table}}

\begin{table}[htb]
\caption{\textbf{Comparison of classification performance (balanced accuracy) of the multi-class approach using different neural network architectures.}}
\label{tab:MC_performance}
\renewcommand{\arraystretch}{1.5}
\begin{tabular}{|l | l | l | l | l |} 
 \hline
  \textbf{Test class} & \makecell{\textbf{EfficientNet}\cite{tan2019efficientnet}, \\ \textbf{pretrained}} & \makecell{\textbf{EfficientNet\cite{tan2019efficientnet}}, \\ \textbf{from scratch}} & \makecell{\textbf{ResNet}\cite{he2016deep}, \\ \textbf{pretrained}} & \makecell{\textbf{LeNet}\cite{lecun1998gradient, krater2021aideveloper}, \\ \textbf{from scratch}} \\ [1ex]
 \hline 
 Basophil & 88.02 $\pm$ 0.94 & \textbf{91.95} $\pm$ 0.89 & 85.07 $\pm$ 2.5 & 88.82 $\pm$ 0.73\\ [0.5ex] 
 \hline
 Eosinophil & \textbf{98.57} $\pm$ 0.05 & 97.25 $\pm$ 0.11 & 97.96 $\pm$ 0.08 & 95.33 $\pm$ 0.19\\ [0.5ex] 
 \hline
 Monocyte & \textbf{98.33} $\pm$ 0.12 & 96.96 $\pm$ 0.23 & 96.71 $\pm$ 0.79 & 94.48 $\pm$ 0.27\\ [0.5ex] 
 \hline
 Neutrophil & \textbf{96.37} $\pm$ 0.39 & 95.88 $\pm$ 0.29 & 95.08 $\pm$ 1.81 &  93.45 $\pm$ 0.38\\ [0.5ex] 
 \hline
 Lymphocyte & \textbf{99.04} $\pm$ 0.17 & 98.43 $\pm$ 0.18 &  98.85 $\pm$ 0.3 & 95.02 $\pm$ 0.13 \\ [0.5ex] 
 \hline
 RBC & \textbf{99.22} $\pm$ 0.02 & 98.88 $\pm$ 0.07 & 99.13 $\pm$ 0.07 & 96.97 $\pm$ 0.33 \\ [0.5ex] 
 \hline
 Nucleated RBC & 96.57 $\pm$ 0.43 & \textbf{97.67} $\pm$ 0.13 & 97.12 $\pm$ 0.55 & 96.28 $\pm$ 0.25\\ [0.5ex]
 \hline
 Platelet & 96.53 $\pm$ 0.28 & 96.1 $\pm$ 0.15 &  95.95 $\pm$ 0.33 & 93.51 $\pm$ 0.59\\ [0.5ex]
 \hline
 RBC aggregate & \textbf{99.81} $\pm$ 0.02 & 99.67 $\pm$ 0.03 &  99.77 $\pm$ 0.05 &  99.29 $\pm$ 0.04\\ [0.5ex]
 \hline
 Platelet aggregate & \textbf{96.04} $\pm$ 0.29 & 94.68 $\pm$ 0.11 &  95.67 $\pm$ 0.1 & 90.96 $\pm$ 0.31 \\ [0.5ex]
 \hline
 RBC \& platelet(s) & \textbf{99.30} $\pm$ 0.07 & 96.7 $\pm$ 0.21 & 98.66 $\pm$ 0.17 & 94.72 $\pm$ 0.43\\ [0.5ex]
 \hline
 WBC \& platelet(s) & \textbf{97.34} $\pm$ 0.09 & 96.6 $\pm$ 0.19 & 95.64 $\pm$ 0.57 & 94.34 $\pm$  0.13\\ [0.5ex]
 \hline
 Average & \textbf{97.1} $\pm$ 0.05 & 96.73 $\pm$ 0.04 &  96.3 $\pm$ 0.12 & 94.43 $\pm$ 0.08\\ [0.5ex]
 \hline
\end{tabular}
\newline
\caption*{Bold numbers indicate the method with higher balanced accuracy. In class names, "(s)" indicates that the class definition allows either a single cell or multiple cells of that type.}
\end{table}

\begin{table}[htb]
\caption{Test set performance (balanced accuracy) of the multi-label classification method on its inherent classes.}
\label{tab:ML_performance_original}
\renewcommand{\arraystretch}{1.5}
\begin{tabular}{|l | l |} 
 \hline
 \textbf{Test class} & \textbf{Performance} \\ [0.5ex] 
 \hline
 Basophil & 84.59 $\pm$ 0.99 \\ [0.5ex] 
 \hline
 Eosinophil & 99.22 $\pm$ 0.03 \\ [0.5ex] 
 \hline
 Monocyte & 98.5 $\pm$ 0.31 \\ [0.5ex] 
 \hline
 Neutrophil & 98.27 $\pm$ 0.18 \\ [0.5ex] 
 \hline
 Lymphocyte & 99.47 $\pm$ 0.06 \\ [0.5ex] 
 \hline
 WBC & 99.45 $\pm$ 0.11 \\ [0.5ex] 
 \hline
 RBC & 99.78 $\pm$ 0.01   \\ [0.5ex] 
 \hline
 Nucleated RBC & 98.42 $\pm$ 0.11 \\ [0.5ex]
 \hline
 Platelet & 99.01 $\pm$ 0.07 \\ [0.5ex]
 \hline
 Generic aggregate & 97.67 $\pm$ 0.06 \\ [0.5ex]
 \hline
 Platelet aggregate & 97.44 $\pm$ 0.11 \\ [0.5ex]
 \hline
 RBC aggregate & 99.79 $\pm$ 0.01 \\ [0.5ex]
 \hline
 Average & 97.63 $\pm$ 0.06 \\ [0.5ex]
 \hline
\end{tabular}
\newline
\caption*{}
\end{table}

\begin{table}[htb]
\caption{Conversion of binary multi-label (ML) outputs to multi-class (MC) outputs for comparison of classification performance.}
\label{tab:ML_to_MC}
{\footnotesize
\renewcommand{\arraystretch}{1.5}
\begin{adjustbox}{max width=\textwidth}
\begin{tabular}{ c  l | c | c | c | c | c | c | c | c | c | c | c | c  } 
\multicolumn{2}{c}{} & \multicolumn{12}{c}{\textbf{Multi-label output}} \\
 & Classes & Baso. & Eosino. & Mono. & Neutro. & Lympho. & WBC & RBC & Nucl. RBC & PLT & Generic aggr. & PLT aggr. & RBC aggr. \\ [0.5ex]   
 \cline{2-14}
 \multirow{12}{*}{\rotatebox{90}{\textbf{Multi-class output}}} & Baso. & \textbf{T} & F & F & F & F & \textbf{T} & F & F & F & F & F & F \\ [0.5ex] 
\cline{2-14}
 & Eosino. & F & \textbf{T} & F & F & F & \textbf{T} & F & F & F & F & F & F \\ [0.5ex]
\cline{2-14}
 & Mono. & F & F & \textbf{T} & F & F & \textbf{T} & F & F & F & F & F & F \\ [0.5ex]
\cline{2-14}
 & Neutro. & F & F & F & \textbf{T} & F & \textbf{T} & F & F & F & F & F & F \\ [0.5ex]
\cline{2-14}
 & Lympho. & F & F & F & F & \textbf{T} & \textbf{T} & F & F & F & F & F & F \\ [0.5ex]
\cline{2-14}
 & RBC & F & F & F & F & F & F & \textbf{T} & F & F & F & F & F \\ [0.5ex]
\cline{2-14}
 & Nucl. RBC & F & F & F & F & F & F & F & \textbf{T} & F & F & F & F \\ [0.5ex]
\cline{2-14}
 & Platelet & F & F & F & F & F & F & F & F & \textbf{T} & F & F & F \\ [0.5ex]
\cline{2-14}
 & RBC aggr. & F & F & F & F & F & F & \textbf{T} & F & F & \textbf{T} & F & \textbf{T} \\ [0.5ex]
\cline{2-14}
 & PLT aggr. & F & F & F & F & F & F & F & F & \textbf{T} & \textbf{T} & \textbf{T} & F \\ [0.5ex]
\cline{2-14}
 & RBC \& PLT(s) & F & F & F & F & F & F & \textbf{T} & F & \textbf{T} & \textbf{T} & - & F \\ [0.5ex]
\cline{2-14}
 & WBC. \& PLT(s) & - & - & - & - & - & \textbf{T} & F & F & \textbf{T} & \textbf{T} & - & F \\ [0.5ex]
\cline{2-14}
& RBC(s) \& PLT aggr. & F & F & F & F & F & F & \textbf{T} & F & \textbf{T} & \textbf{T} & \textbf{T} & - \\ [0.5ex]
\cline{2-14}
& WBC(s) \& PLT aggr. & - & - & - & - & - & \textbf{T} & F & F & \textbf{T} & \textbf{T} & \textbf{T} & F \\ [0.5ex]
\cline{2-14}
& WBC aggr. \& (PLT(s)) & - & - & - & - & - & \textbf{T} & F & F & - & \textbf{T} & - & F \\ [0.5ex]
\cline{2-14}
& RBC(s) \& WBC(s) \& (PLT(s)) & - & - & - & - & - & \textbf{T} & \textbf{T} & F & - & \textbf{T} & - & - \\ [0.5ex]
\end{tabular}
\end{adjustbox}
\newline
\caption*{F: False; T: True; dash: irrelevant. In class names, "(s)" indicates that the class definition allows either a single cell or multiple cells of that type, and names enclosed in brackets indicate that the presence of the corresponding cells is optional. }
}
\end{table}



\end{document}